\documentclass[10pt,twocolumn,letterpaper]{article}

\usepackage{iccv}
\usepackage{times}
\usepackage{epsfig}
\usepackage{graphicx}
\usepackage{amsmath}
\usepackage{amssymb}
\usepackage{mathrsfs}
\usepackage{slashbox}
\usepackage{dsfont}
\usepackage{booktabs}
\usepackage{subcaption}
\usepackage{caption}
\usepackage{nicefrac}
\usepackage{dblfloatfix}
\usepackage{mathrsfs}
\usepackage{xcolor}
\usepackage{colortbl}

\definecolor{sh_gray}{rgb}{0.84,0.84,0.84}
\definecolor{sh_gray2}{rgb}{1,0.89,0.75}
\definecolor{color3}{rgb}{0.95,0.95,0.95}
\definecolor{color4}{rgb}{0.96,0.96,0.86}
\definecolor{color5}{rgb}{0.90,0.90,0.90}

\newlength{\Oldarrayrulewidth}

\usepackage[pagebackref=true,breaklinks=true,letterpaper=true,colorlinks,bookmarks=false]{hyperref}

\iccvfinalcopy 

\def\iccvPaperID{3956} 

\ificcvfinal\fi

\usepackage{cuted}
\usepackage{capt-of}

\begin{document}
\title{Learning Digital Camera Pipeline for Extreme Low-Light Imaging}

\author{Syed Waqas Zamir, Aditya Arora, Salman Khan, Fahad Shahbaz Khan, Ling Shao\\
Inception Institute of Artificial Intelligence, UAE\\
{\tt\small waqas.zamir@inceptioniai.org}
}

 \maketitle

\begin{abstract}\vspace{-0.5em}
In low-light conditions, a conventional camera imaging pipeline produces sub-optimal images that are usually dark and noisy due to a low photon count and low signal-to-noise ratio (SNR). We present a data-driven approach that learns the desired properties of well-exposed images and reflects them in images that are captured in extremely low ambient light environments, thereby significantly improving the visual quality of these low-light images. We propose a new loss function that exploits the characteristics of both pixel-wise and perceptual metrics, enabling our deep neural network to learn the camera processing pipeline to transform the short-exposure, low-light RAW sensor data to well-exposed sRGB images. The results show that our method outperforms the state-of-the-art according to psychophysical tests as well as pixel-wise standard metrics and recent learning-based perceptual image quality measures.
\end{abstract}
\vspace{-0.5cm}

\section{Introduction}
In a dark scene, the ambient light is not sufficient for cameras to accurately capture detail and color information. On one hand, leaving the camera sensor exposed to light for a long period of time retains the actual scene information, but may produce blurred images due to camera shake and object movement in the scene. On the other hand, images taken with a short exposure time preserve sharp details, but are usually dark and noisy. In order to address this dilemma, one might consider taking a sharp picture with a short exposure time and then increasing its brightness. However, the resulting image will not only have amplified noise and blotchy appearance, but the colors will also not match with  those of a corresponding well-exposed image (see, for example, Figure~\hyperlink{fig1b}{1b}). Even if we reduce the problem of noise to some extent by using any state-of-the-art image denoising algorithm, the issue of color remains unsolved \cite{Betalmio2014, Chen2018}.

\begin{figure}[htp]
    \centering
    \begin{subfigure}[t]{0.48\columnwidth}
      \includegraphics[width=\linewidth]{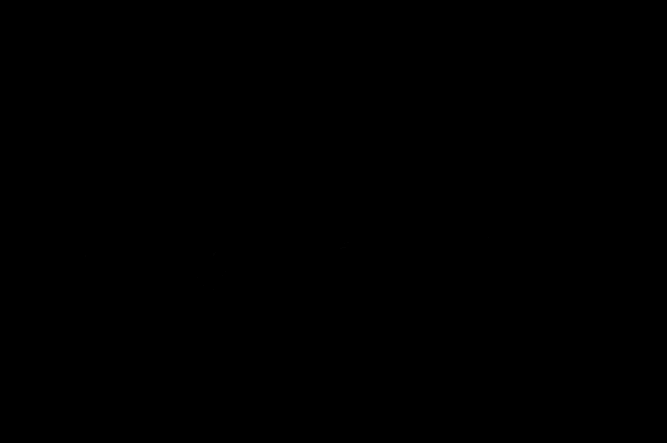}
      \caption{\hypertarget{fig1a}{RAW input image}}
      \label{Fig:teaser_ground_truth}
    \end{subfigure}
    \begin{subfigure}[t]{0.48\columnwidth}
      \includegraphics[width=\linewidth]{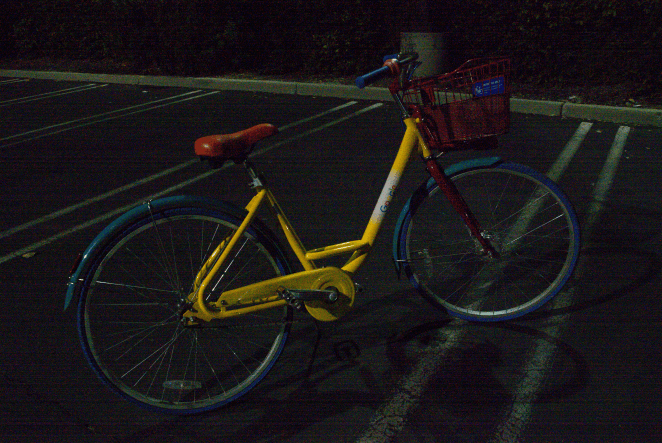}
      \caption{\hypertarget{fig1b}{Traditional pipeline \cite{Ramanath2005}}}
      \label{Fig:teaser_input_image}
    \end{subfigure}
    \begin{subfigure}[t]{0.48\columnwidth}
      \includegraphics[width=\linewidth]{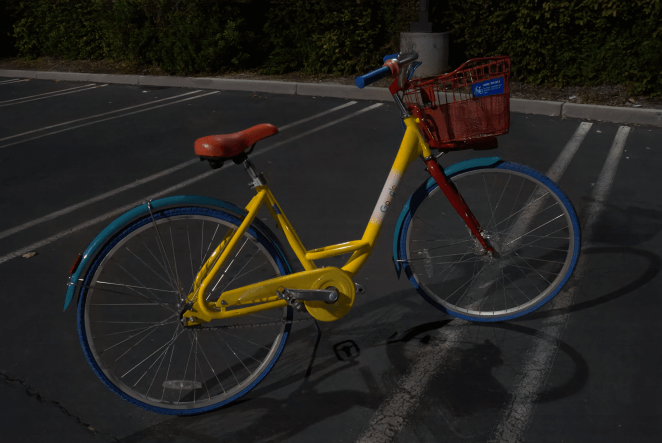}
      \caption{\hypertarget{fig1c}{Chen \etal \cite{Chen2018}}}
      \label{Fig:teaser_chen_et_al}
    \end{subfigure}
    \begin{subfigure}[t]{0.48\columnwidth}
      \includegraphics[width=\linewidth]{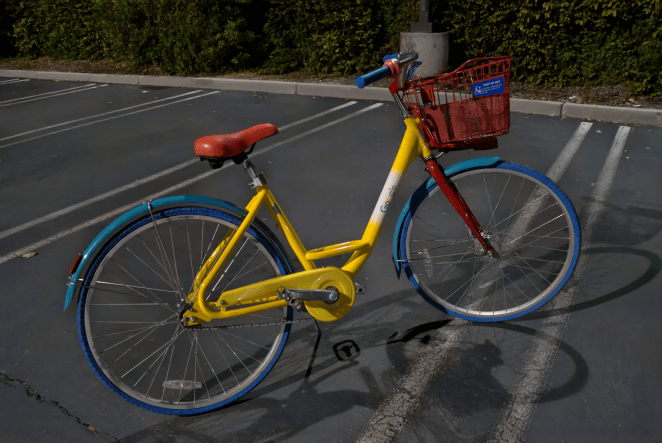}
      \caption{\hypertarget{fig1d}{Our result}}
      \label{Fig:teaser_ours}
    \end{subfigure}
    \captionof{figure}{Transforming a short-exposure RAW image captured in extremely low light to a well-exposed sRGB image. (a) Short-exposed RAW input taken with 0.1s of exposure time. (b) Image produced by traditional camera imaging pipeline \cite{Ramanath2005} applied to input (a). Note that the brightness is increased for better visualization. The reproduced image suffers from noise amplification and a strong color cast. (c) Image produced by the state-of-the-art method \cite{Chen2018}, when applied to (a). (d) Result obtained by our approach, when applied to (a). Compared to \cite{Chen2018}, our method yields an image that is sharper, more vivid, and free from noise and artifacts, while preserving texture and structural information.}
    \label{Fig:teaser}
\end{figure}

A conventional camera imaging pipeline processes the RAW sensor data through a sequence of operations (such as white balance, demosaicking, denoising, color correction, tone mapping, sharpening, etc.)  in order to generate the final RGB images \cite{Ramanath2005}. Solving each of these problems requires hand-crafted priors and, even then, the pipeline breaks down in extremely low-light environments, often yielding dark images with little-to-no visible detail \cite{Chen2018}. 

An alternative way to tackle the issue of low-light imaging is to use deep neural networks. These networks are data hungry in nature and require a large amount of training data: pairs of short-exposure input with corresponding long-exposure ground-truth. To encourage the development of learning-based techniques, Chen \etal \cite{Chen2018} propose a large scale See-in-the-Dark (SID) dataset captured in low light conditions. The SID dataset contains both indoor and outdoor images acquired with two different cameras, having different color filter arrays. They further propose an end-to-end network, employing the $\ell_1$ loss, that learns the complete camera pipeline specifically for low-light imaging. However, the reproduced images often lack contrast and contain artifacts (see Figure~\hyperlink{fig1b}{1c}), especially under extreme low-light environments with severely limited illumination (\eg, dark room with indirect dim light source). 


Most existing image transformation methods \cite{Chen2018, Dong2016,long2015,Zhang2016} focus on measuring the difference between the network's output and the ground-truth, using standard per-pixel loss functions. However, recent studies \cite{Johnson2016,  Prashnani2018, Zhang2018} have shown that applying traditional metrics ($\ell_1$/$\ell_2$,  SSIM \cite{Wang2004}) directly on the pixel-level information often provide overly smooth images that correlate poorly with human perception. These studies, therefore, recommend computing error on the deep feature representations, extracted from any pre-trained network \cite{ He2016,Krizhevsky2012,Szegedy2015}, resulting in images that are visually sharp and perceptually faithful. 
A drawback of such a feature-level error computation strategy is the introduction of checkerboard artifacts at the pixel-level \cite{Johnson2016, Odena2016}. Therefore, information from both \emph{the pixel-level} and \emph{the feature-level} is essential to produce images that are sharp, perceptually faithful and free from artifacts. The aforementioned observation motivates us to develop a new hybrid loss function, exploiting the basic properties of both pixel-wise and perceptual metrics. 

In this paper we propose a data-driven approach based on a novel loss function that is capable of generating well-exposed sRGB images with the desired attributes: sharpness, color vividness, good contrast, noise free, and no color artifacts.
Our end-to-end network takes as input the RAW data captured in extreme low light and generates an sRGB image that fulfills these desired properties. 
By using our new loss function, we learn the entire camera processing pipeline in a supervised manner. Figure~\hyperlink{fig1d}{1d} shows the image produced by the proposed approach. 


\section{Background}
\label{sec:ISP}
Here, we first provide a brief overview of a traditional camera processing pipeline. We then discuss the recently introduced learning-based approach specifically designed for low-light imaging.  

\subsection{Traditional Camera Pipeline}
The basic modules of the imaging pipeline, common to all standard single-sensor cameras, are the following \cite{Ramanath2005}.
\textbf{(a)} \emph{Preprocessing} deals with the issues related to the RAW sensor data such as defective sensor cells, lens shading, light scattering and dark current. 
\textbf{(b)} \emph{White balance} step estimates the scene illumination and remove its effect by linearly scaling the RAW data so that the reproduced image has no color cast \cite{Buchsbaum1980, Karaimer2018}. 
\textbf{(c)} \emph{Demosaicking} stage takes in the RAW data, in which at each pixel location the information of only one color is present, and estimates the other two missing colors by interpolation \cite{Gunturk2005}, yielding a three-channel true color image. 
\textbf{(d)} \emph{Color correction} transforms the image from the sensor-specific color space to linear sRGB color space \cite{Morovic2008}.
\textbf{(e)} \emph{Gamma correction} encodes images by allocating more bits to low luminance values than high luminance values, since we are more perceptible to changes in dark regions than bright areas.
\textbf{(f)} \emph{Post processing} stage applies several camera-specific (proprietary) operations to improve image quality, such as contrast enhancement \cite{Palma-Amestoy2009}, style and aesthetic adjustments \cite{Chen2017, He2018, Yan2016}, denoising \cite{Lefkimmiatis2018, Plotz2017}, and tone mapping \cite{Mantiuk2008}. Optionally, data compression may also be applied \cite{Wallace1991}. 

\begin{figure*}[t]
\begin{center}
       \includegraphics[width=\linewidth]{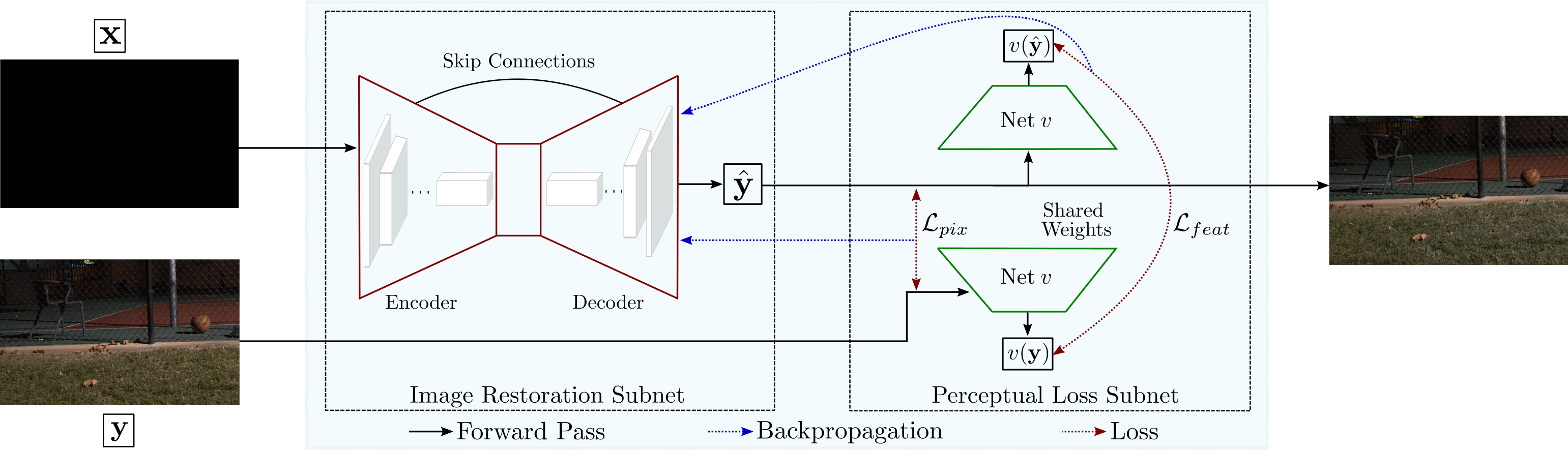}
\end{center}\vspace{-1em}
\caption{Schematic of our framework. Given as input the RAW sensor data $\mathbf{x}$ captured in extremely low ambient light, the image restoration subnet learns the digital camera pipeline and generates a well-exposed sRGB image $\hat{\mathbf{y}}$. The perceptual loss subnet forces the image restoration subnet to produce an output as perceptually similar as possible to the reference image $\mathbf{y}$.}
\label{Fig:network}
\end{figure*}

In low-light environments, the standard camera pipeline provides sub-optimal results due to a low photon count and SNR \cite{Betalmio2014,Chen2018}. 
To acquire well-exposed images in low light, apart from using long exposure, other methods include: exposure bracketing, burst imaging and fusion, larger-aperture lens, flash, and optical image stabilization \cite{Hasinoff2016}. 
However, each of these methods comes with a trade-off and is not always applicable. For instance, a mobile camera has thickness and power constraints, so adding a large lens with fast aperture is infeasible \cite{Hasinoff2016}. In exposure bracketing, a series of images are captured in quick succession with varying shutter speeds and then the user gets to pick the most visually pleasing image from this set, which oftentimes is none of them for difficult lighting. Image fusion for burst imaging often have misalignment problems, leading to ghosting artifacts. Finally, flash photography causes unwanted reflections, glare, shadows, and might change the scene illumination.
In this paper, we address the problem of low light photography using single-imaging systems without flash.

\subsection{Data-driven Image Restoration Approaches}
Deep convolutional neural networks (CNNs) have been used with great success in `\emph{independently}' solving several image processing tasks such as denoising \cite{Lefkimmiatis2018,Plotz2017}, demosaicking \cite{Kokkinos2018}, deblurring \cite{Nah2017, Su2017, Xu2014}, super-resolution \cite{Dong2016,Lai2017,Zhang2018SR}, inpainting \cite{Liu2018,Pathak2016} and contrast enhancement \cite{Gharbi2017,Talebi2018}. 
Recently, learning-based methods \cite{Chen2018, Schwartz2018} have been proposed that `\emph{jointly}' learn the complete camera processing pipeline in an end-to-end manner. Both of these methods take as input the RAW sensor data and produce sRGB images. Particularly, the work of Schwartz \etal \cite{Schwartz2018} deals with images taken in well-lit conditions, and the method of Chen \etal \cite{Chen2018} is developed specifically for extremely low-light imaging. After investigating several loss functions ($\ell_1, \ell_2$, SSIM \cite{Wang2004}, total variation, and GAN \cite{Goodfellow2014}), Chen \etal \cite{Chen2018} opt for a standard pixel-level loss function, i.e., $\ell_1$, to measure the difference between the network's prediction and the ground-truth. However, the per-pixel loss function is restrictive as it only models absolute errors and does not take into account the perceptual quality.  
Next, we propose an approach that exploits the characteristics of both pixel-wise and perceptual metrics to learn the camera processing pipeline in an end-to-end fashion.

\vspace{-0.12em}
\section{Our Method}
Our network design is based on a novel multi-criterion loss formulation, as shown in Figure~\ref{Fig:network}. The model consists of two main blocks: (1) the `\emph{image restoration subnet}', and (2) the `\emph{perceptual loss subnet}'. The image restoration subnet is an encoder-decoder architecture with skip connections between the contraction and expansion pathways. The perceptual loss subnet is a feed-forward CNN.  Here, we first present the loss formulation and later describe each individual block in Sec.~\ref{Sec:network}.

\subsection{Proposed Multi-criterion Loss Function}
\label{sec:proposed loss}
As described earlier, the existing work \cite{Chen2018} for low-light imaging is based on per-pixel loss, i.e., $\ell_1$. We propose a multi-criterion loss function that jointly models the local and global properties of images using pixel-level image details as well as high-level image feature representations. Moreover, it explicitly incorporates perceptual similarity measures to ensure high-quality visual outputs. 

Given an input image  $\mathbf{x}$ and the desired output image $\mathbf{y}$, the image restoration subnet learns a mapping function $f(\mathbf{x};\theta)$. The parameters $\theta$ are updated using the following multi-criterion loss formulation:
\begin{align}\label{eq:our proposed loss}
    \theta^* = \arg\min_{\theta}\mathbb{E}_{\mathbf{x},\mathbf{y}}\Big[\sum\limits_{k}\alpha_k \mathcal{L}_k(g^k(\mathbf{x};\psi), h^k(\mathbf{y}; \phi))\Big] 
\end{align}
where, $\mathcal{L}_k$ denotes the individual loss function, and $g^k(\cdot), h^k(\cdot)$ are functions on the input and target image, respectively, whose definitions vary depending on the type of loss. In this paper, we consider two distinct representation levels (pixel-level and feature-level) to compute two loss criterion, i.e., $\mathcal{L}_k \in \{\mathcal{L}_{pix}, \mathcal{L}_{feat}\}$. The first loss criterion, $\mathcal{L}_{pix}$, is pixel-based and accounts for low-level image detail. The pixel-level loss is further divided into two terms: standard $\ell_1$ loss and structure similarity loss. The second loss criterion, $\mathcal{L}_{feat}$, is a high-level perceptual loss based on intermediate deep feature representations.  
Next, we elaborate on these pixel-level and feature-level error criterion.

\subsubsection{Pixel Loss: $\mathcal{L}_{pix}$}
The $\mathcal{L}_{pix}$ loss in Eq. (\ref{eq:our proposed loss}) computes error directly on the pixel-level information of the network's output and the ground-truth image. In this case, the definitions of $g^{pix}$ and $h^{pix}$ are fairly straight-forward:
$g^{pix} = f(\mathbf{x}; \theta) = \hat{\mathbf{y}}, \; h^{pix} = \mathds{1}(\mathbf{y}).$
The loss function is defined as:
\begin{align}
\mathcal{L}_{pix} = \beta \ell_1(\hat{\mathbf{y}},\mathbf{y}) + (1-\beta) \mathcal{L}_{\text{MS-SSIM}}(\hat{\mathbf{y}},\mathbf{y})
\label{Eq:pixel loss}
\end{align}
where $\beta \in [0,1]$ is a weight parameter that we set using grid search on the validation set.
\vspace{0.3em}\\ 
\textbf{Absolute deviation.} 
The $\ell_1$ error directly minimizes the difference between the network output and the ground-truth to transform low-light images to well-exposed ones. Given $\hat{\mathbf{y}}$ and $\mathbf{y}$, the $\ell_1$ loss can be computed as:
\begin{equation}
\ell_1(\hat{\mathbf{y}} ,\mathbf{y}) = \frac{1}{N}\sum_{p=1}^{N}\lvert \hat{\mathbf{y}}_p - \mathbf{y}_p  \rvert,
\end{equation} 
where $p$ is the pixel location and $N$ denotes the total number of pixels in the image.

Although the $\ell_1$ metric is a popular choice for the loss function, it compromises high-frequency details, such as texture and sharp edges. 
To avoid such artifacts, we introduce a structural similarity measure in Eq. (\ref{Eq:pixel loss}).
\vspace{0.30em}\\
\textbf{Structural similarity measure.} 
This term ensures the perceptual change in the structural content of output images to be minimal. In this work, we utilize the multi-scale structural similarity measure (MS-SSIM) \cite{Wang2003}:
\begin{equation}
\mathcal{L}_{\text{MS-SSIM}}(\hat{\mathbf{y}},\mathbf{y}) = 1- \frac{1}{N}\sum_{p=1}^{N} \text{MS-SSIM}(\hat{\mathbf{y}}_p,   \mathbf{y}_p).
\end{equation}

In order to define $\text{MS-SSIM}$, let us assume $\mu_{\hat{y}}$, $\sigma_{\hat{y}}^2$ and $\sigma_{{\hat{y}}y}$ are the mean of image $
\hat{\mathbf{y}}$, the variance of $\hat{\mathbf{y}}$, and the covariance of image $\hat{\mathbf{y}}$ and image $\mathbf{y}$, respectively. Then,
\begin{equation}
\text{SSIM}(\hat{\mathbf{y}},\mathbf{y}) = \frac{2 \mu_{\hat{\mathbf{y}}} \mu_\mathbf{y} + C_1}{\mu_{\hat{\mathbf{y}}}^2 + \mu_\mathbf{y}^2 + C_1} \cdot \frac{2\sigma_{\hat{\mathbf{y}}\mathbf{y}} + C_2}{\sigma_{\hat{\mathbf{y}}}^2+\sigma_\mathbf{y}^2 + C_2}
\label{Eq:SSIM}
\end{equation}
\begin{equation}
  = l(\hat{\mathbf{y}},\mathbf{y}) \cdot cs(\hat{\mathbf{y}},\mathbf{y})
\end{equation}
and finally, 
\begin{equation}
\text{MS-SSIM}(\hat{\mathbf{y}},\mathbf{y}) = \left[ l_M(\hat{\mathbf{y}},\mathbf{y})\right]^{\gamma_M} \cdot \prod_{i=1}^{M}\left[cs_i(\hat{\mathbf{y}},\mathbf{y})\right]^{\eta_i}, 
\label{Eq:MSSSIM}
\end{equation}
where, $M$ is the number of scales. The first term in Eq. (\ref{Eq:MSSSIM}) compares the luminance of image $\hat{\mathbf{y}}$ with the luminance of reference image $\mathbf{y}$, and it is computed only at scale $M$. The second term measures the contrast and structural differences at various scales. $\gamma_M$ and $\eta_i$ adjust the relative importance of each term and, for convenience, we set $\gamma_M = \eta_i = 1$ for $i = \{1,...,M\}$. $C_1$ and $C_2$ in Eq. (\ref{Eq:SSIM}) are small constants \cite{Wang2003}.

\subsubsection{Feature Loss: $\mathcal{L}_{feat}$}
The pixel-level loss term is valuable for preserving original colors and detail in the reproduced images. However, it does not integrate perceptually sound global scene detail since the structural similarity is only enforced locally. To resolve this problem, we propose to use an additional loss term that quantifies the perceptual viability of the generated outputs in terms of a higher-order feature representation obtained from the perceptual loss subnet (see Figure~\ref{Fig:network}).

In the feature loss term of the objective function (\ref{eq:our proposed loss}), instead of calculating errors directly on the pixel-level, we measure the difference between the feature representations of the output and ground-truth images extracted with a deep network \cite{Simonyan2015} pre-trained on the ImageNet dataset \cite{Deng2009}. Note that this choice is motivated from a recent large-scale study \cite{Zhang2018} that demonstrates the suitability of deep features as a perceptual metric. In this case, the functions $g^{feat}$ and $h^{feat}$ are defined as $g^{feat} = h^{feat} = v^l(\cdot)$, where $v^l(\cdot)$ denotes the $l^{th}$ layer activation map from the the network. The loss term is formulated as:
\begin{equation}
\mathcal{L}_{feat}(\hat{\mathbf{y}},\mathbf{y}) = \frac{1}{N}\parallel v^l(\hat{\mathbf{y}}; \psi) - v^l(\mathbf{y}; \psi)  \parallel_2^2,
\label{Eq:feature loss}
\end{equation}
In this work, we use the VGG-16 network \cite{Simonyan2015}. Note that other image classification networks such as AlexNet \cite{Krizhevsky2012}, ResNet \cite{He2016}, or GoogLeNet \cite{Szegedy2015} can also be used to extract feature representations \cite{Zhang2018}. The perceptual loss function $\mathcal{L}_{feat}$ \cite{Johnson2016} enforces our \emph{image restoration subnet} to generate outputs that are perceptually similar to their corresponding well-exposed reference images. 



\subsection{Network Architecture}
\label{Sec:network}
Here we provide details of both blocks of our framework (see Figure~\ref{Fig:network}).
\vspace{0.30em}\\
\textbf{Image restoration subnet.} Our network inherits a U-net encoder-decoder structure \cite{ronneberger2015u} with symmetric skip connections between the lower layers of the encoder and the corresponding higher layers of the decoder. The benefits of such a design for the restoration subnet are three-fold: (a) it has superior performance on image restoration and segmentation tasks \cite{Chen2018,ronneberger2015u}, (b) it can process a full-resolution image (i.e., at $4240{\times}2832$ or $6000{\times}4000$ resolution) due to its fully convolutional design and low memory signature, and (c) the skip connections between the encoder and decoder modules enable adequate propagation of context information and preserve high-resolution details. Our network operates on RAW sensor data rather than RGB images, since our objective is to replace the traditional camera pipeline with an automatically learned network.

The image restoration subnet consists of a total of 23 convolutional layers. Among these, the \emph{encoder} module has 10 convolutional layers, arranged as five pairs of $3{\times}3$ layers. Each pair is followed by a leaky ReLU non-linearity $\left(LReLU(x) = \text{max}(0,x) + 0.2\text{min}(0,x)\right)$ and a $2{\times}2$ max-pooling operator for subsampling. The \emph{decoder} module has a total of 13 convolutional layers. These layers are arranged as a set of four blocks, each of which consists of a transpose convolutional layer whose output is concatenated with the corresponding features maps from the encoder module, followed by two $3{\times}3$ convolutional layers. The number of channels in the feature maps are progressively reduced  and the spatial resolution is increased due to the transpose convolutional layers. Finally, a $1{\times}1$ convolutional layer, followed by a sub-pixel layer \cite{Shi2016}, is applied to remap the channels and obtain the RGB image with the same spatial resolution as the original RAW image. (For more details on network design and for a toy example, see supplementary material.) 
\vspace{0.35em}\\
\textbf{Perceptual loss subnet.}
The perceptual loss subnet consists of a truncated version of VGG-16 \cite{Simonyan2015}. We only use the first two convolutional layers of VGG-16 and obtain the feature representation after ReLU non-linearity. This feature representation has been demonstrated to accurately encode the style and perceptual content of an image \cite{Johnson2016}. The result is a $\nicefrac{H}{4} \times \nicefrac{W}{4} \times 128$ tensor for both the output of the image restoration net and the ground-truth, which are then used to compute the similarity between them. 
\section{Experiments}

\subsection{Dataset}
\begin{figure}[t]
\begin{center}
       \includegraphics[width=\linewidth]{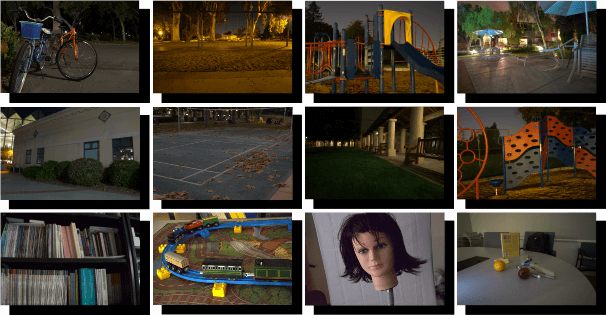}
\end{center}\vspace{-1.4em}
\caption{Some sample images from the See-in-the-Dark (SID) dataset \cite{Chen2018}: long-exposure ground truth images (in front), and short-exposure and essentially black input images (in background). Note that the reference images in the last row are noisy, indicating the presence of a very high noise level in their corresponding short-exposure input images, thus making the problem even more challenging.}
\label{Fig:dataset}
\end{figure}

We validate our approach on the See-in-the-Dark (SID) dataset \cite{Chen2018} that was specifically collected for the development of learning-based methods for low-light photography. In Figure~\ref{Fig:dataset} we show some sample images from the SID dataset. 
The images were captured using two different cameras: Sony $\alpha$7S II with a Bayer color filter array (CFA) and sensor resolution of 4240$\times$2832, and Fujifilm X-T2 with a X-Trans CFA and 6000$\times$4000 spatial resolution.  
The dataset contains 5094 short-exposure RAW input images and their corresponding long-exposure reference images. Note that there are 424 unique long-exposure reference images, indicating that multiple short-exposure input images can correspond to the same ground-truth image.
There are both indoor and outdoor images of the static scenes. The ambient illuminance reaching the camera was in the range 0.2 to 5 lux for outdoor scenes and between 0.03 lux and 0.3 lux for indoor scenes. Input images were taken with an exposure time between 1/30 and 1/10 seconds and the exposure time for the ground-truth images was 10 to 30 seconds.

\subsection{Camera-specific Preprocessing}
\label{Sec:preprocessing}
As mentioned in Sec.~\ref{sec:ISP}, cameras have a CFA in front of the image sensor to capture color information. Different cameras use different types of CFAs; Bayer filter array being the most popular choice due to its simple layout. 
Images of the SID dataset \cite{Chen2018} come from cameras with different CFAs. Therefore, before passing the RAW input to the \emph{image restoration subnet} (Figure~\ref{Fig:network}), we pack the data, as in \cite{Chen2018}, into $4$ channels if it comes from Bayer CFA and $9$ channels for X-Trans CFA.

At the borders of the image sensor, there are some pixels that never see the light and therefore should be zero (black). However, during image acquisition, the values of these pixels are raised due to thermally generated voltage. We subtract this camera-specific black level from the image signal.
Finally, we scale the sensor data with an amplification factor (\eg, $\times$100, $\times$250, or $\times$300), which is the ratio between the reference image and the input image and determines the brightness of the output image.

\begin{figure*}[t]
\begin{center}
    \begin{subfigure}[t]{0.33\textwidth}
      \includegraphics[width=\linewidth]{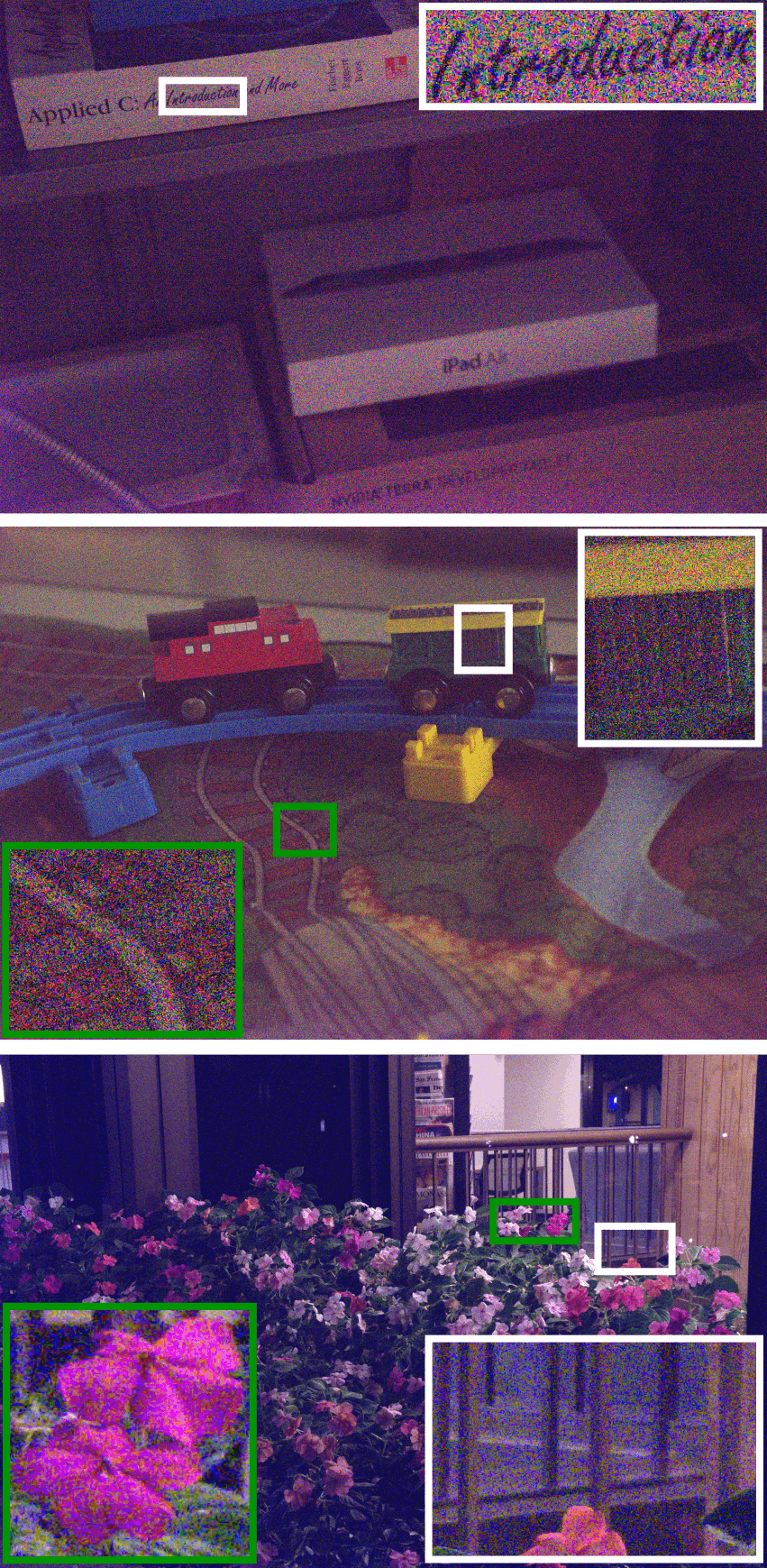}
      \caption{Traditional pipeline}
      \label{Fig:results traditional}
    \end{subfigure}
    \begin{subfigure}[t]{0.33\textwidth}
      \includegraphics[width=\linewidth]{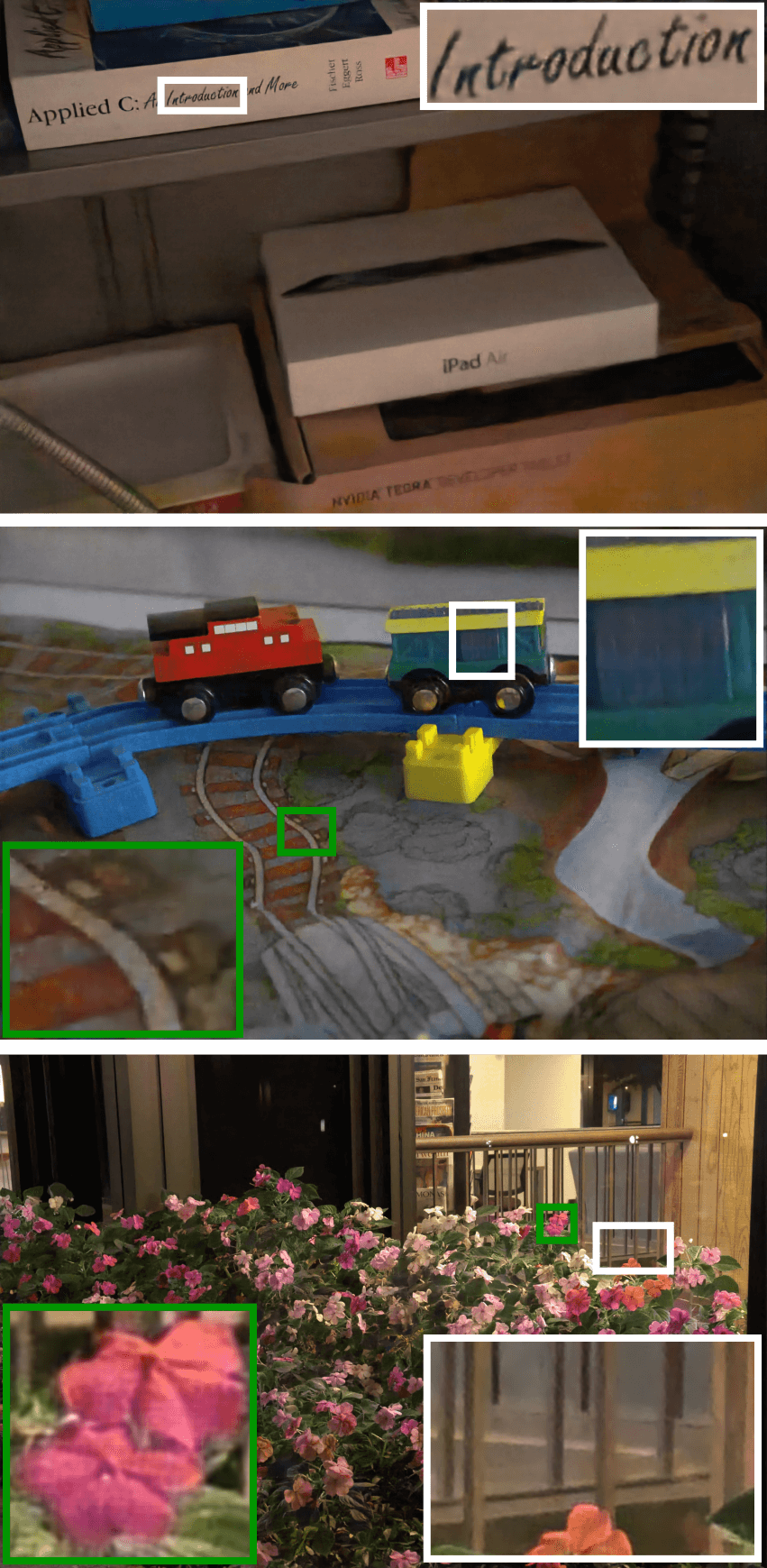}
      \caption{Chen \etal. \cite{Chen2018}}
      \label{Fig:results chen}
    \end{subfigure}
    \begin{subfigure}[t]{0.33\textwidth}
      \includegraphics[width=\linewidth]{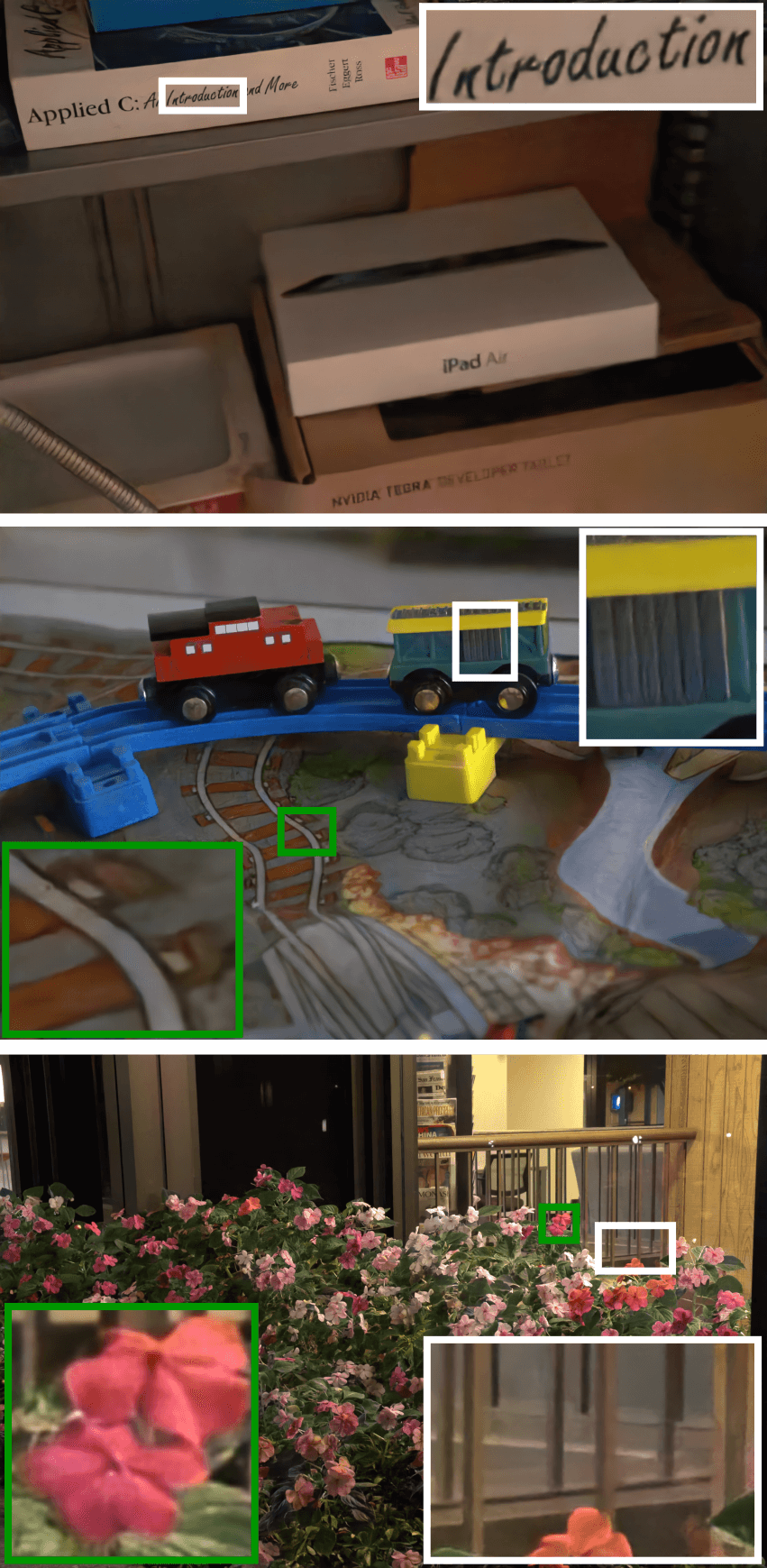}
      \caption{Our results}
      \label{Fig:results ours}
    \end{subfigure}
\end{center}\vspace{-1em}
\caption{Qualitative comparison of our approach with the state-of-the-art method \cite{Chen2018} and the traditional pipeline. (a) Images produced by the conventional pipeline are noisy and contain strong color artifacts. (b) The approach of \cite{Chen2018} generates images with splotchy textures, color distortions and poorly reconstructed shapes (zoomed-in regions). (c) Our method produces images that are sharp, colorful and noise free. }
\label{Fig:results}
\end{figure*}

\begin{figure}[t]
\begin{center}
    \begin{subfigure}[t]{0.235\textwidth}
      \includegraphics[width=\linewidth]{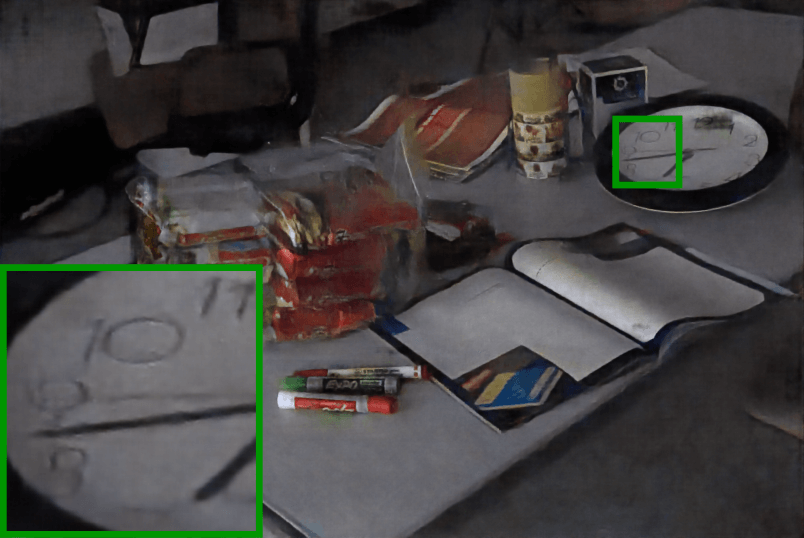}
      \caption{Chen \etal. \cite{Chen2018}}
      \label{Fig:results chen extreme}
    \end{subfigure}
    \begin{subfigure}[t]{0.235\textwidth}
      \includegraphics[width=\linewidth]{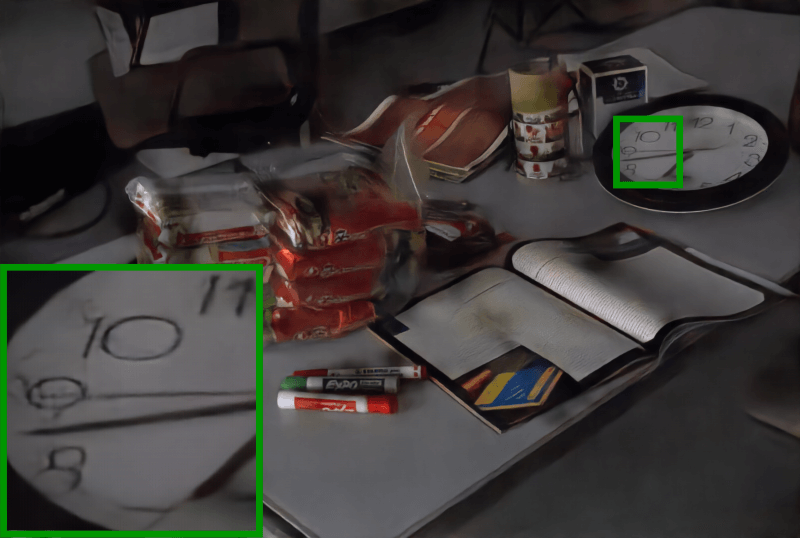}
      \caption{Our result}
      \label{Fig:results ours extreme}
    \end{subfigure}
\end{center}\vspace{-1em}
\caption{Visual example in extremely difficult lighting. Compare the (zoomed-in) clock and the other objects.}
\label{Fig:results extreme}
\end{figure}

\subsection{Training}
We train two separate networks: one for the Sony subset and the other for the Fuji subset from the SID dataset \cite{Chen2018}. Each network takes as input a short-exposure RAW image and a corresponding long-exposure reference image (which is converted into the sRGB color space with the \emph{LibRAW} library). 
Note that the input is prepared using camera-specific preprocessing mentioned in Sec.~\ref{Sec:preprocessing}, before being passed through our network (Figure~\ref{Fig:network}). Both networks are trained for $4000$ epochs using the proposed loss function (Sec.~\ref{sec:proposed loss}). We use Adam optimizer with an initial learning rate of $10^{-4}$, which is reduced to $10^{-5}$ after $2000$ epochs. In each iteration we take a $512 \times 512$ crop from the training image and perform random rotation and flipping. To compute the $\mathcal{L}_{feat}$ loss (\ref{Eq:feature loss}), we use features from the $conv2$ layer after ReLU of the VGG-16 network. 
The batch size is set to one, as we observed that setting the batch size greater than one reduces accuracy. This might be because the network struggles to learn, at once, the transformation process for images having significantly different light and noise levels. 
We empirically set $\alpha = 0.9$ and $\beta = 0.99$ in Eq. (\ref{eq:our proposed loss}) and Eq. (\ref{Eq:pixel loss}), respectively, for all the experiments.

\subsection{Qualitative Evaluation}
To the best of our knowledge, Chen \etal \cite{Chen2018} present the ``\emph{first and only}'' data-driven work that learns the digital camera pipeline specifically for extreme low-light imaging.  
Figure \ref{Fig:results} presents a qualitative comparison of the images produced by our method and those of the state-of-the-art technique \cite{Chen2018}, as well as the traditional camera processing pipeline.
Note that the traditional pipeline provides dark images with little-to-no visible detail. Therefore, we scale the brightness of the results of the traditional pipeline for visualization purposes.
It is apparent in Figure~\ref{Fig:results traditional} that the traditional pipeline handles low-light images poorly and yields results with extreme noise, color cast and artifacts.
As reported in \cite{Betalmio2014,Chen2018}, applying a state-of-the-art image denoising algorithm \cite{Guo2018,Lefkimmiatis2018,Plotz2017} might reduce noise to some extent. However, the issue of color distortion remains unsolved.

Figure  \ref{Fig:results} further shows that the results produced by our model are noticeably sharper, better denoised, more natural and visually pleasant, compared to those generated by the state-of-the-art method \cite{Chen2018}. For instance, it can be seen in Figure~\ref{Fig:results chen} that the image reproductions of \cite{Chen2018} exhibit splotchy textures (text, rail-track), color distortions (train, flowers, bottom corners of row 1) and poorly reconstructed shapes, such as for the text, rail-track and wooden fence. 
\vspace{0.35em}\\
\textbf{Extremely challenging case.} 
In Figure~\ref{Fig:results extreme}, we show the performance of our method on an (example) image captured in extremely difficult lighting: dark room with indirect dim light source. Our result might not be acceptable in isolation, but when compared with \cite{Chen2018}, we can greatly appreciate the reconstruction of sharp edges such as for the digits of the clock, and the spatial smoothness of the homogeneous regions, such as the table top and the floor. 

\begin{table*}[!htp]
\begin{center}
\setlength{\tabcolsep}{14pt}
\begin{tabular}{l l c c p{3mm} c c}
\toprule
\rowcolor{color3} & & \multicolumn{2}{c}{Experienced Observers} & & \multicolumn{2}{c}{Inexperienced Observers} \\
\cline{3-4} \cline{6-7}
 \rowcolor{color3} & & x100 Set & x250 Set & & x100 Set & x250 Set\\
\midrule
\multicolumn{1}{l}{Sony Dataset}  & Ours $>$ Chen \etal \cite{Chen2018} & 84.7\% & 92.6\% & & 80.3\% & 86.6\%
\\
\multicolumn{1}{l}{Fuji Dataset}  & Ours $>$ Chen \etal \cite{Chen2018}  & 76.1\% & 89.7\% & & 80.9\% & 89.2\%
\\
\bottomrule
\end{tabular}
\end{center}\vspace{-0.5em}
\caption{Psychophysical experiments: 7 expert and 18 naive observers compare the results produced by our method and Chen \etal \cite{Chen2018}. Our method significantly outperforms \cite{Chen2018} in both the easier x100 and the challenging x250 test images.}
\label{Table:results sup}
\end{table*}

\begin{table*}[!htp]
\begin{center}
\setlength{\tabcolsep}{8.7pt}
\begin{tabular}{l c c c p{3mm} c c c}
\toprule
\rowcolor{color3} & \multicolumn{3}{c}{Sony subset \cite{Chen2018}} & & \multicolumn{3}{c}{Fuji subset \cite{Chen2018}} \\
\cline{2-4} \cline{6-8}
 \rowcolor{color3}& PSNR $\textcolor{red}{\uparrow}$& PieAPP \cite{Prashnani2018} $\textcolor{red}{\downarrow}$ & LPIPS \cite{Zhang2018} $\textcolor{red}{\downarrow}$  & & PSNR $\textcolor{red}{\uparrow}$ & PieAPP \cite{Prashnani2018} $\textcolor{red}{\downarrow}$ & LPIPS \cite{Zhang2018} $\textcolor{red}{\downarrow}$  \\
\midrule
\multicolumn{1}{l}{Chen \textit{et al.} \cite{Chen2018}} & 
29.18  & 1.576  & 0.470 & &
 27.34 & 1.957  & 0.598 \\
\multicolumn{1}{l}{Ours} & 
\textbf{29.43} & \textbf{1.511} &  \textbf{0.443} & & 
\textbf{27.63} & \textbf{1.763} & \textbf{0.476}\\
\bottomrule
\end{tabular}
\end{center}\vspace{-0.5em}
\caption{Quantitative comparison using four full-reference metrics on the  SID dataset. The results are reported as mean errors. Our method provides superior performance compared to the state-of-the-art  \cite{Chen2018}. $\textcolor{red}{\downarrow}$: lower is better. $\textcolor{red}{\uparrow}$: higher is better.}
\label{Table:results}
\end{table*}

\begin{table*}[!htp]
\begin{center}
\setlength{\tabcolsep}{8pt}
\begin{tabular}{l c c c c c c c}
\toprule
\rowcolor{color3} & $\ell_1$ (Chen \etal \cite{Chen2018})  & MS-SSIM & $\mathcal{L}_{pix}$ & $\mathcal{L}_{feat}$ & $\mathcal{L}_{feat}$ + $\ell_1$ & $\mathcal{L}_{feat}$ + $\mathcal{L}_{\text{MS-SSIM}}$ & $\mathcal{L}_{final}$ \\
\midrule
\multicolumn{1}{l}{Sony subset \cite{Chen2018}} & 29.18 & 29.37 & 29.33 & 27.34 &  29.22 & 29.40 & \textbf{29.43} \\
\multicolumn{1}{l}{Fuji subset \cite{Chen2018}} &  27.34 & 27.55 & 27.51 & 23.07 &  27.37 & 27.52 & \textbf{27.63} \\
\bottomrule
\end{tabular}
\end{center}\vspace{-0.5em}
\caption{Ablation study: impact of each individual term of the proposed loss function on the final results. Each term contributes to the overall performance. Results are reported on the test images of SID dataset in terms of mean PSNR.}
\label{Table:ablation study}
\end{table*}

\subsection{Subjective Evaluation of Perceptual Quality} We conduct a psychophysical experiment to assess the performance of competing approaches in an office-like environment. A corpus of 25 observers with normal color vision participated in the experiment. The subjects belong to two different groups: 7 expert observers with prior experience in image processing, and 18 naive observers. Each observer was shown a pair of corresponding images on the screen, sequentially and in random order: one of these images is produced by our method and the other one by Chen \etal \cite{Chen2018}. Observers were asked to examine the color, texture, structure, sharpness and artifacts, and then choose the image which they find more pleasant. Each participant repeated this process on the test images of the Sony and Fuji subsets from the SID dataset \cite{Chen2018}.
The percentage with which the observers preferred images produced by our method than those of Chen \etal \cite{Chen2018} is reported in Table \ref{Table:results sup}. These results indicate that our method significantly outperforms the state-of-the-art \cite{Chen2018} in terms of perceptual quality.

\vspace{-0.12em}
\subsection{Quantitative Evaluation}
To perform a quantitative assessment of the results, we use two recent learning-based perceptual metrics (LPIPS \cite{Zhang2018} and PieAPP \cite{Prashnani2018}) and the standard PSNR metric. For the sake of fair comparison, we leave SSIM metric \cite{Wang2004} out from the evaluation as our method is optimized using its variant MS-SSIM \cite{Wang2003}.
The average values of these metrics for the testing images of the Sony and Fuji subsets \cite{Chen2018} are reported in Table \ref{Table:results}.  
Our method outperforms the state-of-the-art \cite{Chen2018} by a considerable margin.
\vspace{0.4em}\\
\textbf{Ablation study.}
The proposed loss function that minimizes the error of the network consists of three individual terms ($\ell_1$, $\mathcal{L}_{\text{MS-SSIM}}$ and $\mathcal{L}_{feat}$). Here, we evaluate the impact of each individual term and their combinations on our end-task. Table \ref{Table:ablation study} summarizes our results where we compare different loss variants using the exact same parameter ($\alpha, \beta$) settings. Our results demonstrate that each individual term contributes towards the final performance of our method. Based on the PSNR values in Table \ref{Table:ablation study} and our qualitative observations, we draw the following conclusions: \textbf{(a)}
each individual component has its respective limitations e.g., $\ell_1$ yields colorful results but with artifacts, $\mathcal{L}_{\text{MS-SSIM}}$ preserves fine image details but provides less saturated results, $\mathcal{L}_{feat}$ reconstructs structure well, but introduces checkerboard artifacts. \textbf{(b)} The combination of $\ell_1$, $\mathcal{L}_{\text{MS-SSIM}}$ and $\mathcal{L}_{feat}$ in an appropriate proportion provides the best results. The final loss function accumulates the complementary strengths of each individual criterion and avoids their respective shortcomings. The resulting images are colorful and artifact free, while faithfully preserving image structure and texture \footnote{Additional results are provided in supplementary material.}.




\begin{figure*}[t]
\begin{center}
      \begin{subfigure}[t]{0.245\textwidth}
      \includegraphics[width=\linewidth]{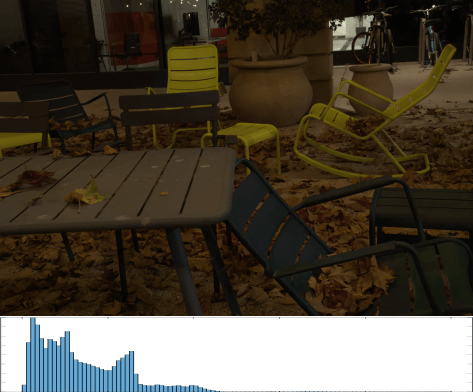}
      \caption{}
      \label{Fig:dehazing fig a}
    \end{subfigure}
    \begin{subfigure}[t]{0.245\textwidth}
      \includegraphics[width=\linewidth]{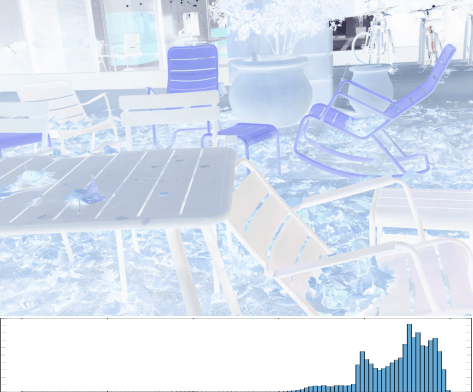}
      \caption{}
      \label{Fig:dehazing fig b}
    \end{subfigure}
    \begin{subfigure}[t]{0.245\textwidth}
      \includegraphics[width=\linewidth]{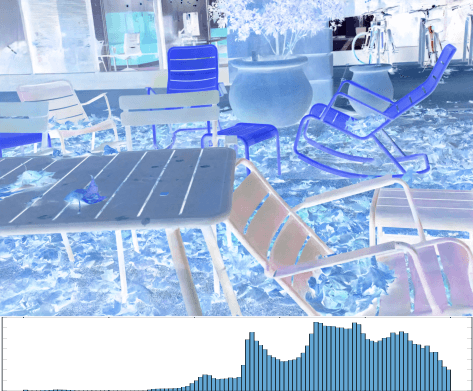}
      \caption{}
      \label{Fig:dehazing fig c}
    \end{subfigure}
    \begin{subfigure}[t]{0.245\textwidth}
      \includegraphics[width=\linewidth]{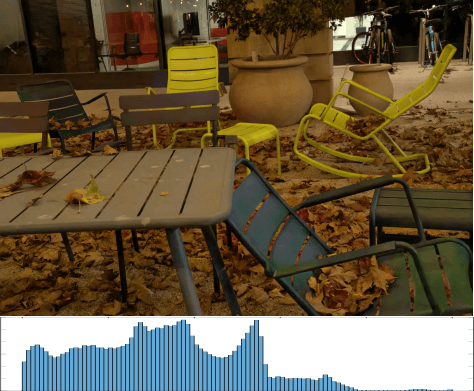}
      \caption{}
      \label{Fig:dehazing fig d}
    \end{subfigure}
\end{center}\vspace{-0.5em}
\caption{Effect of contrast improvement procedure. (a) Output of our \emph{image restoration subnet} and its histogram. (b) Inverting the image of (a). (c) Output obtained after applying image dehazing algorithm \cite{He2011} on image from (b). (d) Final enhanced image obtained by inverting (c). Note that histograms are computed using the lightness component of the images.}
\label{Fig:dehazing}
\end{figure*}

\begin{figure}[t]
\begin{center}
    \begin{subfigure}[t]{0.235\textwidth}
      \includegraphics[width=\linewidth]{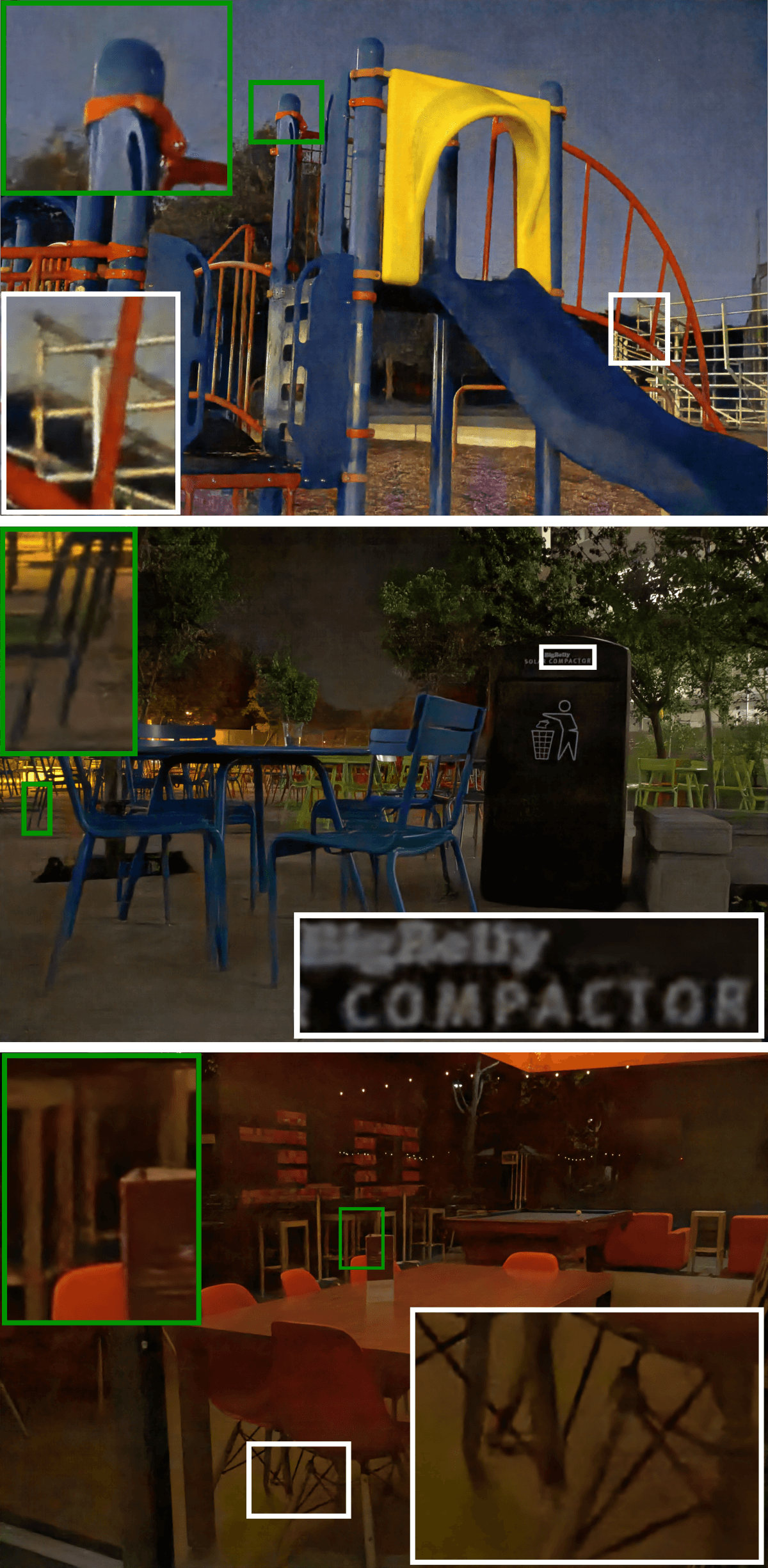}
      \caption{Chen \etal. \cite{Chen2018}}
      \label{Fig:results chen dehazed}
    \end{subfigure}
    \begin{subfigure}[t]{0.235\textwidth}
      \includegraphics[width=\linewidth]{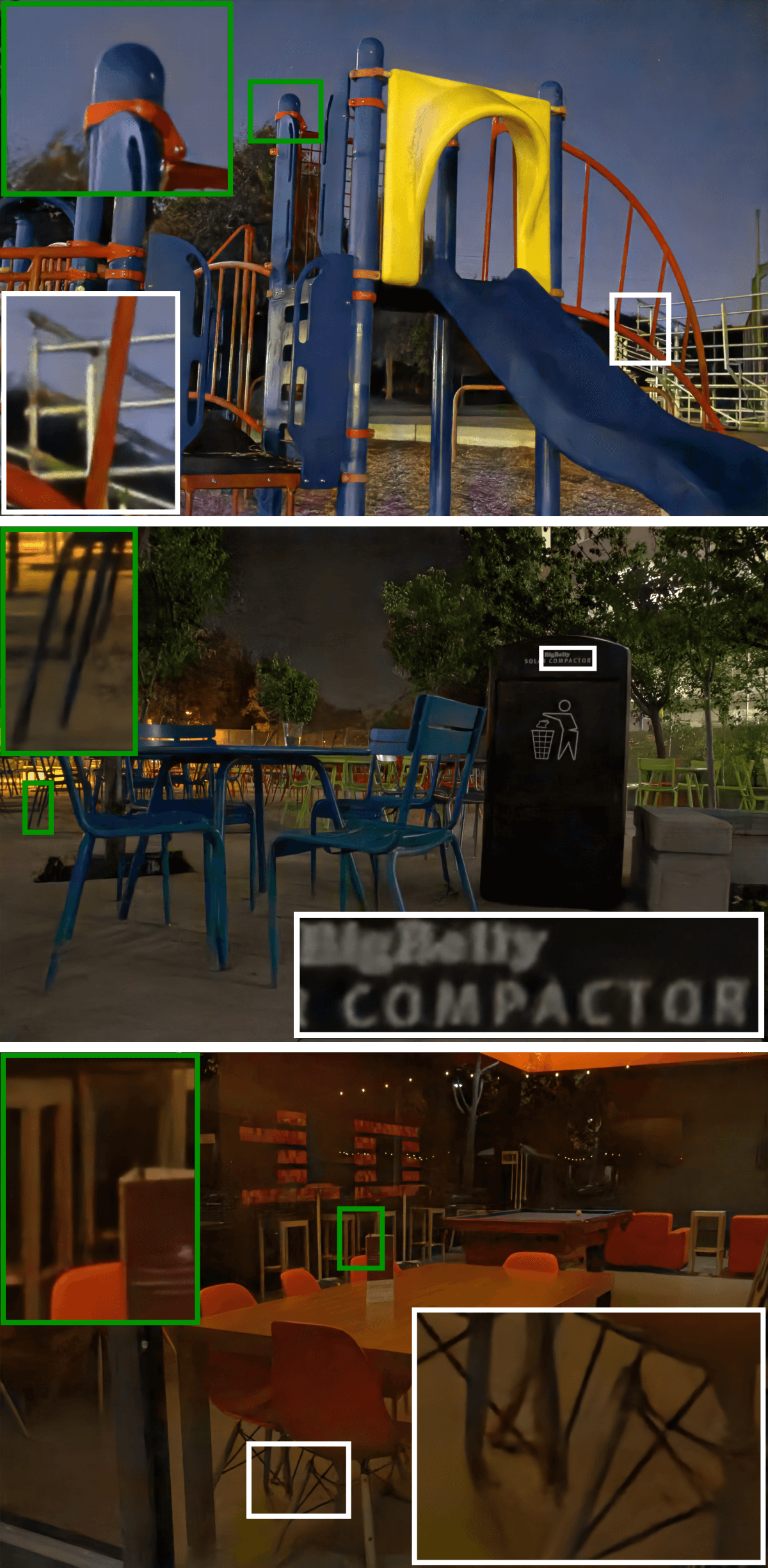}
      \caption{Our results}
      \label{Fig:results ours dehazed}
    \end{subfigure}
\end{center}\vspace{-0.5em}
\caption{Effect of contrast contrast improvement procedure when applied on our results and those of \cite{Chen2018}. The visual quality of our results is further improved. Whereas, in the results of \cite{Chen2018} the artifacts become even more apparent.}
\label{Fig:results dehazed}
\end{figure}

\section{Contrast Improvement Procedure}
The network of Chen \etal \cite{Chen2018} produces images that are often dark and lack contrast. It is the inherent limitation enforced by the \emph{imperfect} ground-truth of the SID dataset, and therefore learned network will also be only partially optimal. In attempt to dealing with this issue, \cite{Chen2018} preprocesses the ground-truth images with histogram equalization. Subsequently, their network learns to generate contrast enhanced outputs; however, with artifacts. Thus the performance of the method \cite{Chen2018} was significantly reduced.


Inspired from \cite{Dong2011}, we employ the following procedure in order to improve the color contrast of the results produced by our proposed method.
We observe that the histogram of outputs produced by our image restoration subnet is mostly skewed towards dark regions (see for example Figure~\ref{Fig:dehazing fig a}). 
By inverting the intensity values of the image, the histogram becomes similar to that of a hazy image (Figure~\ref{Fig:dehazing fig b}). This indicates that, by applying an image dehazing algorithm \cite{He2011}, we can make the image histogram more uniform (Figure~\ref{Fig:dehazing fig c}). Finally, inverting back the intensities of the image provides us with a new image that is bright, sharp, colorful and without artifacts, as shown in Figure~\ref{Fig:dehazing fig d}. 

We notice that preprocessing the reference images by applying the just mentioned procedure and then training the network from scratch produces suboptimal results. Therefore, we first train the network with regular ground-truth for 4000 epochs, and then perform fine-tuning for another 100 epochs with the contrast-enhanced ground-truth. 

In Figure~\ref{Fig:results dehazed}, we compare the results obtained after applying the contrast improvement strategy to our image restoration net and to the framework of Chen \etal \cite{Chen2018}. It is evident that our method produces visually compelling images with good contrast and vivid colors. Whereas the method of \cite{Chen2018} has a tendency of reproducing images with artifacts, which become even more prominent when the contrast enhancement procedure is employed; notice the zoomed-in portions of Figure~\ref{Fig:results dehazed}, especially the sky in column 1.

\section{Conclusion}
Imaging in extremely low-light conditions is a highly challenging task for the conventional camera pipeline, often yielding dark and noisy images with little-to-no detail. In this paper, we proposed a learning-based approach that learns the entire camera pipeline, end-to-end, for low-light conditions. 
We explored the benefits of computing loss both at the pixel-level information and at the feature-level, and presented a new loss function that significantly improved the performance.
We conducted a psychophysical study, according to which the observers overwhelmingly preferred the outputs of our method over the existing state-of-the-art. Similar trends were observed when the image quality of the competing methods was assessed with standard metrics, as well as recent learning-based error metrics.

{\small
\bibliographystyle{ieee}
\bibliography{bib}

\begin{thebibliography}{10}\itemsep=-1pt

\bibitem{Betalmio2014}
M.~Bertalm\'io and S.~Levine.
\newblock Variational approach for the fusion of exposure bracketed pairs.
\newblock {\em TIP}, 22(2):712--723, 2013.

\bibitem{Buchsbaum1980}
G.~Buchsbaum.
\newblock A spatial processor model for object colour perception.
\newblock {\em Journal of the Franklin Institute}, 310(1):1 -- 26, 1980.

\bibitem{Chen2018}
C.~Chen, Q.~Chen, J.~Xu, and V.~Koltun.
\newblock Learning to see in the dark.
\newblock In {\em CVPR}, 2018.

\bibitem{Chen2017}
Q.~Chen, J.~Xu, and V.~Koltun.
\newblock Fast image processing with fully-convolutional networks.
\newblock In {\em ICCV}, 2017.

\bibitem{Deng2009}
J.~Deng, W.~Dong, R.~Socher, L.~Li, K.~Li, and L.~Fei-Fei.
\newblock {ImageNet}: A large-scale hierarchical image database.
\newblock In {\em CVPR}, 2009.

\bibitem{Dong2016}
C.~Dong, C.~C. Loy, K.~He, and X.~Tang.
\newblock Image super-resolution using deep convolutional networks.
\newblock {\em TPAMI}, 38(2):295--307, 2016.

\bibitem{Dong2011}
X.~Dong, G.~Wang, Y.~Pang, W.~Li, M.~W. Wen, J., and Y.~Lu.
\newblock Fast efficient algorithm for enhancement of low lighting video.
\newblock In {\em ICME}, pages 1--6, 2011.

\bibitem{Gharbi2017}
M.~Gharbi, J.~Chen, J.~T. Barron, S.~W. Hasinoff, and F.~Durand.
\newblock Deep bilateral learning for real-time image enhancement.
\newblock {\em TOG}, 36(4):118, 2017.

\bibitem{Goodfellow2014}
I.~Goodfellow, J.~Pouget-Abadie, M.~Mirza, B.~Xu, D.~Warde-Farley, S.~Ozair,
  A.~Courville, and Y.~Bengio.
\newblock Generative adversarial nets.
\newblock In {\em NIPs}, 2014.

\bibitem{Gunturk2005}
B.~K. Gunturk, J.~Glotzbach, Y.~Altunbasak, R.~W. Schafer, and R.~M. Mersereau.
\newblock Demosaicking: color filter array interpolation.
\newblock {\em IEEE Signal Processing Magazine}, 22(1):44--54, 2005.

\bibitem{Guo2018}
S.~Guo, Z.~Yan, K.~Zhang, W.~Zuo, and L.~Zhang.
\newblock Toward convolutional blind denoising of real photographs.
\newblock {\em arXiv preprint arXiv:1807.04686}, 2018.

\bibitem{Hasinoff2016}
S.~W. Hasinoff, D.~Sharlet, R.~Geiss, A.~Adams, J.~T. Barron, F.~Kainz,
  J.~Chen, and M.~Levoy.
\newblock Burst photography for high dynamic range and low-light imaging on
  mobile cameras.
\newblock {\em TOG}, 35(6):192, 2016.

\bibitem{He2011}
K.~He, J.~Sun, and X.~Tang.
\newblock Single image haze removal using dark channel prior.
\newblock {\em TPAMI}, 33(12):2341--2353, 2011.

\bibitem{He2016}
K.~He, X.~Zhang, S.~Ren, and J.~Sun.
\newblock Deep residual learning for image recognition.
\newblock In {\em CVPR}, 2016.

\bibitem{He2018}
M.~He, D.~Chen, J.~Liao, P.~V. Sander, and L.~Yuan.
\newblock Deep exemplar-based colorization.
\newblock {\em TOG}, 37(4):47, 2018.

\bibitem{Johnson2016}
J.~Johnson, A.~Alahi, and L.~Fei-Fei.
\newblock Perceptual losses for real-time style transfer and super-resolution.
\newblock In {\em ECCV}, 2016.

\bibitem{Karaimer2018}
H.~C. Karaimer and M.~S. Brown.
\newblock Improving color reproduction accuracy on cameras.
\newblock In {\em CVPR}, 2018.

\bibitem{Kokkinos2018}
F.~Kokkinos and S.~Lefkimmiatis.
\newblock Deep image demosaicking using a cascade of convolutional residual
  denoising networks.
\newblock In {\em ECCV}, 2018.

\bibitem{Krizhevsky2012}
A.~Krizhevsky, I.~Sutskever, and G.~E. Hinton.
\newblock {ImageNet} classification with deep convolutional neural networks.
\newblock In {\em NIPS}, 2012.

\bibitem{Lai2017}
W.-S. Lai, J.-B. Huang, N.~Ahuja, and M.-H. Yang.
\newblock Deep laplacian pyramid networks for fast and accurate
  superresolution.
\newblock In {\em CVPR}, 2017.

\bibitem{Lefkimmiatis2018}
S.~Lefkimmiatis.
\newblock Universal denoising networks: A novel {CNN} architecture for image
  denoising.
\newblock In {\em CVPR}, 2018.

\bibitem{Liu2018}
G.~Liu, F.~A. Reda, K.~J. Shih, T.-C. Wang, A.~Tao, and B.~Catanzaro.
\newblock Image inpainting for irregular holes using partial convolutions.
\newblock In {\em ECCV}, 2018.

\bibitem{long2015}
J.~Long, E.~Shelhamer, and T.~Darrell.
\newblock Fully convolutional networks for semantic segmentation.
\newblock In {\em CVPR}, 2015.

\bibitem{Mantiuk2008}
R.~Mantiuk, S.~Daly, and L.~Kerofsky.
\newblock Display adaptive tone mapping.
\newblock {\em TOG}, 27(3):1--10, 2008.

\bibitem{Morovic2008}
J.~Morovi{\v{c}}.
\newblock {\em Color gamut mapping}, volume~10.
\newblock Wiley, 2008.

\bibitem{Nah2017}
S.~Nah, T.~H. Kim, and K.~M. Lee.
\newblock Deep multi-scale convolutional neural network for dynamic scene
  deblurring.
\newblock In {\em CVPR}, 2017.

\bibitem{Odena2016}
A.~Odena, V.~Dumoulin, and C.~Olah.
\newblock Deconvolution and checkerboard artifacts.
\newblock {\em Distill}, 2016.

\bibitem{Palma-Amestoy2009}
R.~Palma-Amestoy, E.~Provenzi, M.~Bertalm\'io, and V.~Caselles.
\newblock A perceptually inspired variational framework for color enhancement.
\newblock {\em TPAMI}, 31(3):458--474, 2009.

\bibitem{Pathak2016}
D.~Pathak, P.~Krahenbuhl, J.~Donahue, T.~Darrell, and A.~A. Efros.
\newblock Context encoders: Feature learning by inpainting.
\newblock In {\em CVPR}, 2016.

\bibitem{Plotz2017}
T.~Plotz and S.~Roth.
\newblock Benchmarking denoising algorithms with real photographs.
\newblock In {\em CVPR}, 2017.

\bibitem{Prashnani2018}
E.~Prashnani, H.~Cai, Y.~Mostofi, and P.~Sen.
\newblock {PieAPP}: Perceptual image-error assessment through pairwise
  preference.
\newblock In {\em CVPR}, 2018.

\bibitem{Ramanath2005}
R.~Ramanath, W.~E. Snyder, Y.~Yoo, and M.~S. Drew.
\newblock Color image processing pipeline.
\newblock {\em IEEE Signal Processing Magazine}, 22(1):34--43, 2005.

\bibitem{ronneberger2015u}
O.~Ronneberger, P.~Fischer, and T.~Brox.
\newblock U-net: Convolutional networks for biomedical image segmentation.
\newblock In {\em MICCAI}, 2015.

\bibitem{Schwartz2018}
E.~Schwartz, R.~Giryes, and A.~M. Bronstein.
\newblock {DeepISP}: Towards learning an end-to-end image processing pipeline.
\newblock {\em TIP}, 2018.
\newblock (Early Access).

\bibitem{Shi2016}
W.~Shi, J.~Caballero, F.~Husz{\'a}r, J.~Totz, A.~P. Aitken, R.~Bishop,
  D.~Rueckert, and Z.~Wang.
\newblock Real-time single image and video super-resolution using an efficient
  sub-pixel convolutional neural network.
\newblock In {\em CVPR}, 2016.

\bibitem{Simonyan2015}
K.~Simonyan and A.~Zisserman.
\newblock Very deep convolutional networks for large-scale image recognition.
\newblock In {\em ICLR}, 2015.

\bibitem{Su2017}
S.~Su, M.~Delbracio, J.~Wang, G.~Sapiro, W.~Heidrich, and O.~Wang.
\newblock Deep video deblurring for hand-held cameras.
\newblock In {\em CVPR}, 2017.

\bibitem{Szegedy2015}
C.~Szegedy, W.~Liu, Y.~Jia, P.~Sermanet, S.~Reed, D.~Anguelov, D.~Erhan,
  V.~Vanhoucke, and A.~Rabinovich.
\newblock Going deeper with convolutions.
\newblock In {\em CVPR}, 2015.

\bibitem{Talebi2018}
H.~Talebi and P.~Milanfar.
\newblock Learned perceptual image enhancement.
\newblock In {\em ICCP}, 2018.

\bibitem{Wallace1991}
G.~K. Wallace.
\newblock The {JPEG} still picture compression standard.
\newblock {\em ACM - Special issue on digital multimedia Commun.},
  34(4):30--44, 1991.

\bibitem{Wang2004}
Z.~Wang, A.~C. Bovik, H.~R. Sheikh, and E.~P. Simoncelli.
\newblock Image quality assessment: from error visibility to structural
  similarity.
\newblock {\em TIP}, 13(4):600--612, 2004.

\bibitem{Wang2003}
Z.~Wang, E.~P. Simoncelli, and A.~C. Bovik.
\newblock Multiscale structural similarity for image quality assessment.
\newblock In {\em ACSSC}, 2003.

\bibitem{Xu2014}
L.~Xu, J.~S.~J. Ren, C.~Liu, and J.~Jia.
\newblock Deep convolutional neural network for image deconvolution.
\newblock In {\em NIPS}, 2014.

\bibitem{Yan2016}
Z.~Yan, H.~Zhang, B.~Wang, S.~Paris, and Y.~Yu.
\newblock Automatic photo adjustment using deep neural networks.
\newblock {\em TOG}, 35(2):1--15, 2016.

\bibitem{Zhang2016}
R.~Zhang, P.~Isola, and A.~A. Efros.
\newblock Colorful image colorization.
\newblock In {\em ECCV}, 2016.

\bibitem{Zhang2018}
R.~Zhang, P.~Isola, A.~A. Efros, E.~Shechtman, and O.~Wang.
\newblock The unreasonable effectiveness of deep features as a perceptual
  metric.
\newblock In {\em CVPR}, 2018.

\bibitem{Zhang2018SR}
Y.~Zhang, K.~Li, K.~Li, L.~Wang, B.~Zhong, and Y.~Fu.
\newblock Image super-resolution using very deep residual channel attention
  networks.
\newblock In {\em ECCV}, 2018.

\end{thebibliography}
}

\end{document}